\documentclass[runningheads]{llncs}
\usepackage{graphicx}
\usepackage{comment}
\usepackage{amsmath,amssymb} 
\usepackage{color}
\usepackage{booktabs}
\usepackage{multirow}
\usepackage{microtype}
\usepackage{subfig}

\usepackage[pagebackref=true,breaklinks=true,letterpaper=true,colorlinks,bookmarks=false]{hyperref}

\begin{document}
\pagestyle{headings}
\mainmatter
\def\ECCVSubNumber{513}  

\title{Unified Multisensory Perception: Weakly-Supervised Audio-Visual Video Parsing} 

\titlerunning{Weakly-Supervised Audio-Visual Video Parsing}
%
\author{Yapeng Tian\inst{1}\and
Dingzeyu Li\inst{2} \and
Chenliang Xu\inst{1}}
\authorrunning{Y. Tian et al.}
%
\institute{University of Rochester \and
Adobe Research\\
\email{\{yapengtian,chenliang.xu\}@rochester.edu}; \email{dinli@adobe.com}}
\maketitle

\begin{abstract}
In this paper, we introduce a new problem, named audio-visual video parsing, which aims to parse a video into temporal event segments and label them as either audible, visible, or both. Such a problem is essential for a complete understanding of the scene depicted inside a video. To facilitate exploration, we collect a \textit{Look, Listen, and Parse} (LLP) dataset to investigate audio-visual video parsing in a weakly-supervised manner. This task can be naturally formulated as a Multimodal Multiple Instance Learning (MMIL) problem. Concretely, we propose a novel hybrid attention network to explore unimodal and cross-modal temporal contexts simultaneously. We develop an attentive MMIL pooling method to adaptively explore useful audio and visual content from different temporal extent and modalities. Furthermore, we discover and mitigate modality bias and noisy label issues with an individual-guided learning mechanism and label smoothing technique, respectively. Experimental results show that the challenging audio-visual video parsing can be achieved even with only video-level weak labels. Our proposed framework can effectively leverage unimodal and cross-modal temporal contexts and alleviate modality bias and noisy labels problems. 

\end{abstract}

\section{Introduction}
\label{sec:intro}

\begin{figure}[tb]
    \centering
    \includegraphics[width=\columnwidth]{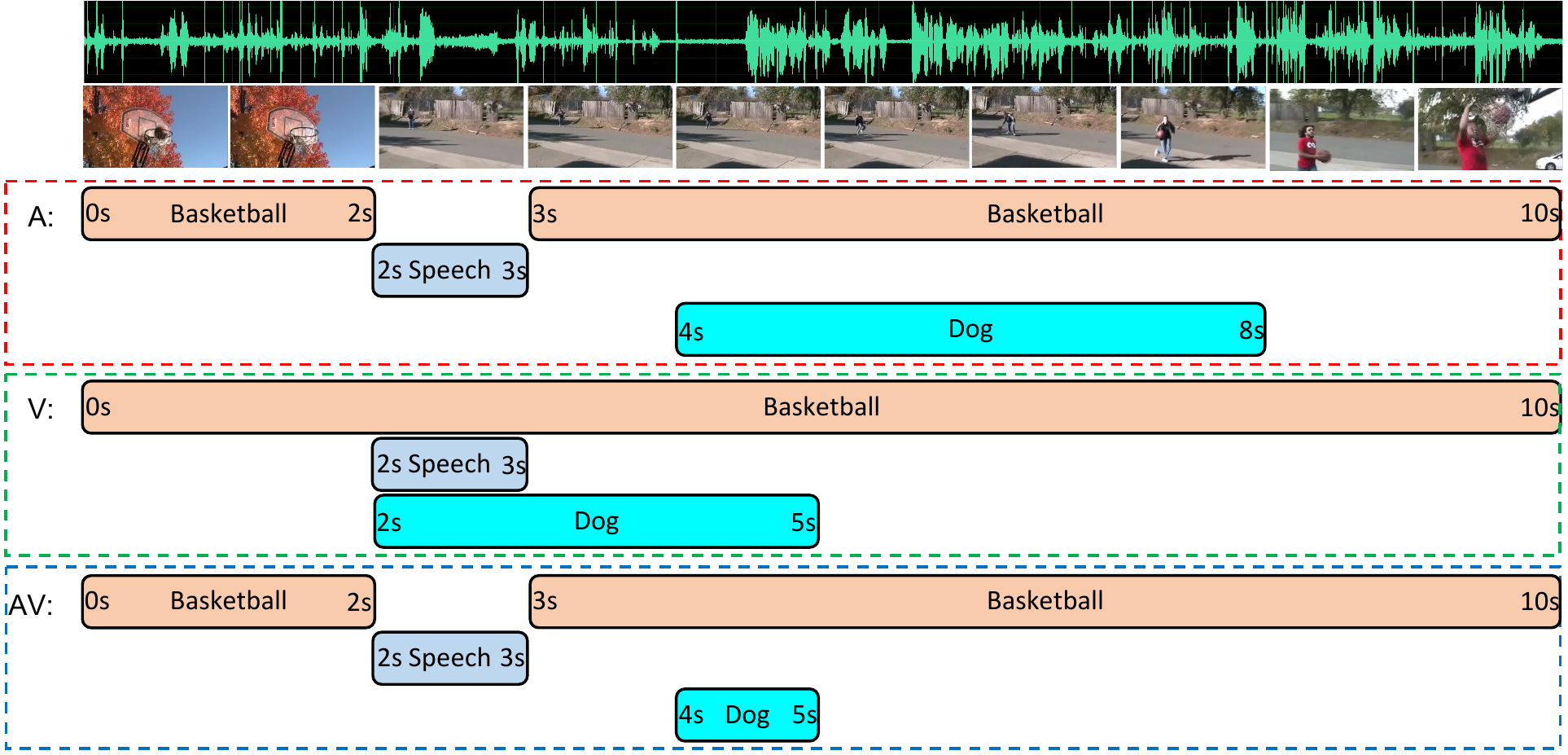}
    \caption{Our audio-visual video parsing model aims to parse a video into different audio (audible), visual (visible), and audio-visual (audi-visible) events with correct categories and boundaries. A dog in the video visually appears from 2nd second to 5th second and make barking sounds from 4th second to 8th second. So, we have audio event (4s-8s), visual event (2s-5s), and audio-visual event (4s-5s) for the \textit{Dog} event category.}
    \label{fig:teaser}
\end{figure} 


Human perception involves complex analyses of visual, auditory, tactile, gustatory, olfactory, and other sensory data. 
Numerous psychological and brain cognitive studies~\cite{bulkin2006seeing,jacobs2019can,shams2008benefits,spence2003multisensory} show that combining different sensory data is crucial for human perception. 
However, the vast majority of work~\cite{gaidon2013temporal,lin2014microsoft,shou2016temporal,yang2009video} in scene understanding, an essential perception task, focuses on visual-only methods ignoring other sensory modalities. 
They are inherently limited. For example, when the object of interest is outside of the field-of-view (FoV), one would rely on audio cues for localization. 
While there is little data on tactile, gustatory, or olfactory signals, we do have an abundance of multimodal audiovisual data, e.g., YouTube videos.


Utilizing and learning from both auditory and visual modalities is an emerging research topic. Recent years have seen progress in learning representations~\cite{Arandjelovic2017ICCV,aytar2016soundnet,hu2019deep,korbar2018cooperative,owens2018audio,owens2016ambient}, separating visually indicated sounds~\cite{ephrat2018looking,gao2018learning,gao20192,gao2019co,zhao2019sound,zhao2018sound,gan2020music,zhou2020sep}, spatially localizing visible sound sources~\cite{owens2018audio,senocak2018learning,tian2018audio}, and temporally localizing audio-visual synchronized segments~\cite{lin2019dual,tian2018audio,wu2019DAM}. However, past approaches usually assume audio and visual data are always correlated or even temporally aligned. In practice, when we analyze the video scene, many videos have audible sounds, which originate outside of the FoV, leaving no visual correspondences, but still contribute to the overall understanding, such as out-of-screen running cars and a narrating person.
Such examples are ubiquitous, which leads us to some basic questions: what video events are audible, visible, and ``audi-visible,'' where and when are these events inside of a video, and how can we effectively detect them? 


To answer the above questions, we pose and try to tackle a fundamental problem: \textit{audio-visual video parsing} that recognizes event categories bind to sensory modalities, and meanwhile, finds temporal boundaries of when such an event starts and ends (see Fig.~\ref{fig:teaser}).
However, learning a fully supervised audio-visual video parsing model requires densely annotated event modality and category labels with corresponding event onsets and offsets, which will make the labeling process extremely expensive and time-consuming. 
To avoid tedious labeling, we explore weakly-supervised learning for the task, which only requires sparse labeling on the presence or absence of video events. 
The weak labels are easier to annotate and can be gathered in a large scale from web videos.

We formulate the weakly-supervised audio-visual video parsing as a Multimodal Multiple Instance Learning (MMIL) problem and propose a new framework to solve it. 
Concretely, we use a new hybrid attention network (HAN) for leveraging unimodal and cross-modal temporal contexts simultaneously. We develop an attentive MMIL pooling method for adaptively aggregating useful audio and visual content from different temporal extent and modalities. Furthermore, we discover modality bias and noisy label issues and alleviate them with an individual-guided learning mechanism and label smoothing~\cite{reed2014training}, respectively. 

To facilitate our investigations, we collect a \textit{Look, listen, and Parse} (LLP) dataset that has $11,849$ YouTube video clips from $25$ event categories. We label them with sparse video-level event labels for training. 
For evaluation, we label a set of precise labels, including event modalities, event categories, and their temporal boundaries. Experimental results show that it is tractable to learn audio-visual video parsing even with video-level weak labels. Our proposed HAN model can effectively leverage multimodal temporal contexts. Furthermore, modality bias and noisy label problems can be addressed with the proposed individual learning strategy and label smoothing, respectively. Besides, we make a discussion on the potential applications enabled by audio-visual video parsing. 

The contributions of our work include: (1) a new audio-visual video parsing task towards a unified multisensory perception; (2) a novel hybrid attention network to leverage unimodal and cross-modal temporal contexts simultaneously; (3) an effective attentive MMIL pooling to aggregate multimodal information adaptively; (4) a new individual guided learning approach to mitigate the modality bias in the MMIL problem and label smoothing to alleviate noisy labels; and (5) a newly collected large-scale video dataset, named LLP, for audio-visual video parsing. Dataset, code, and pre-trained models are publicly available in \url{https://github.com/YapengTian/AVVP-ECCV20}.

\section{Related Work}
\label{sec:related}

In this section, we discuss some related work on temporal action localization, sound event detection, and audio-visual learning.

\noindent\textbf{Temporal Action Localization.} Temporal action localization (TAL) methods usually use sliding windows as action candidates and address TAL as a classification problem~\cite{gaidon2013temporal,lin2018bsn,long2019gaussian,shou2017cdc,shou2016temporal,zhao2017temporal} learning from full supervisions. Recently, weakly-supervised approaches are proposed to solve the TAL. Wang \emph{et al.}~\cite{wang2017untrimmednets} present an UntrimmedNet with a classification module and a selection module to learn the action models and reason about the
temporal duration of action instances, respectively. Hide-and-seek~\cite{singh2017hide} randomly hides certain sequences while training to force the model to explore more discriminative content. Paul~\emph{et al.}~\cite{paul2018w} introduce a co-activity similarity loss to enforce instances in the same class to be similar in the feature space. Inspired by the class activation map method~\cite{zhou2016learning}, Nguyen~\emph{et al.}~\cite{nguyen2018weakly} propose a sparse temporal pooling network (STPN). Liu~\emph{et al.}~\cite{liu2019completeness} incorporate both action completeness modeling and action-context separation into a weakly-supervised TAL framework. Unlike actions in TAL, video events in audio-visual video parsing might contain motionless or even out-of-screen sound sources and the events can be perceived by either audio or visual modalities. Even though, we extend two recent weakly-supervised TAL methods: STPN~\cite{nguyen2018weakly} and CMCS~\cite{liu2019completeness} to address visual event parsing and compare them with our model in Sec.~\ref{sec:comp}. 

\noindent\textbf{Sound Event Detection.}
Sound event detection (SED) is a task of recognizing and locating audio events in acoustic environments. Early supervised approaches rely on some machine learning models, such as support vector machines~\cite{elizalde2016experiments}, Gaussian mixture models~\cite{heittola2013context} and recurrent neural networks~\cite{parascandolo2016recurrent}. To bypass strongly labeled data, weakly-supervised SED methods are developed~\cite{chou2018learning,kong2018audio,mcfee2018adaptive,wang2019comparison}. These methods only focus on audio events from constrained domains, such as urban sounds~\cite{salamon2014dataset} and domestic environments~\cite{mesaros2017dcase} and visual information is ignored. However, our audio-visual video parsing will exploit both modalities to parse not only event categories and boundaries but also event perceiving modalities towards a unified multisensory perception for unconstrained videos.

\noindent\textbf{Audio-Visual Learning.} Benefiting from the natural synchronization between auditory and visual modalities, audio-visual learning has enabled a set of new problems and applications including representation learning~\cite{Arandjelovic2017ICCV,aytar2016soundnet,hu2019deep,korbar2018cooperative,ngiam2011multimodal,owens2018audio,owens2016ambient}, audio-visual sound separation~\cite{ephrat2018looking,gao2018learning,gao20192,gao2019co,zhao2019sound,zhao2018sound,gan2020music,zhou2020sep}, vision-infused audio inpainting~\cite{zhou2019vision}, sound source spatial localization~\cite{owens2018audio,senocak2018learning,tian2018audio}, sound-assisted action recognition~\cite{gao2019listentolook,kazakos2019epic,korbar2019scsampler}, audio-visual video captioning~\cite{rahman2019watch,tian2018attempt,Tian_2019_CVPR_Workshops,wang2018watch}, and audio-visual event localization~\cite{lin2019dual,tian2018audio,tian2019audio,wu2019DAM}. Most previous work assumes that temporally synchronized audio and visual content are always matched conveying the same semantic meanings. However, unconstrained videos can be very noisy: sound sources might not be visible (\emph{e.g.}, an out-of-screen running car and a narrating person) and not all visible objects are audible (\emph{e.g.}, a static motorcycle and people dancing with music).
Different from previous methods, we pose and seek to tackle a fundamental but unexplored problem: audio-visual video parsing for parsing unconstrained videos into a set of video events associated with event categories, boundaries, and modalities. Since the existing methods cannot directly address our problem, we modify the recent weakly-supervised audio-visual event 
localization methods: AVE~\cite{tian2018audio} and AVSDN~\cite{lin2019dual} adding additional audio and visual parsing branches as baselines.



\begin{figure}[tb]
    \centering
    \includegraphics[width=\columnwidth]{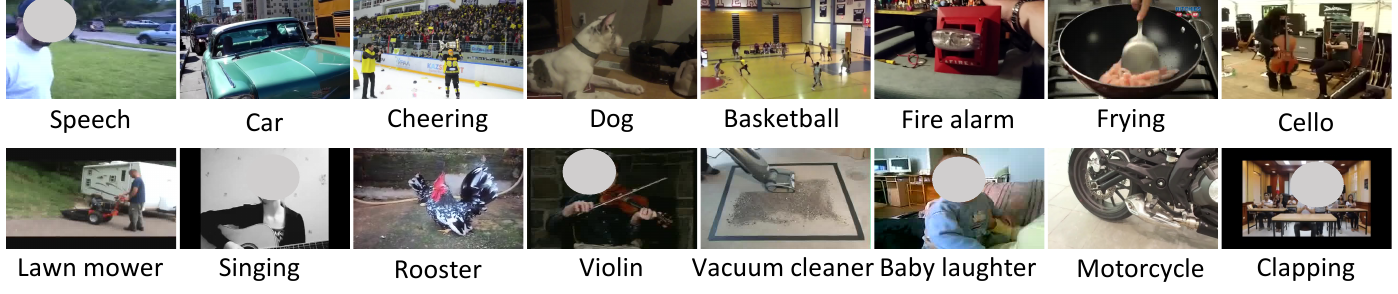}
    \caption{Some examples from the LLP dataset.}
    \label{fig:data}
\end{figure} 

\section{LLP: The Look, Listen and Parse Dataset} 

To the best of our knowledge, there is no existing dataset that is suitable for our research. 
Thus, we introduce a \textit{Look, Listen, and Parse} dataset for audio-visual video scene parsing, which contains 11,849 YouTube video clips spanning over 25 categories for
a total of 32.9 hours collected from AudioSet~\cite{gemmeke2017audio}.
A wide range of video events (\emph{e.g.}, human speaking, singing, baby crying, dog barking, violin playing, and car running, and vacuum cleaning \emph{etc}.) from diverse domains (\emph{e.g.}, human activities, animal activities, music performances, vehicle sounds, and domestic environments) are included in the dataset. Some examples in the LLP dataset are shown in Fig.~\ref{fig:data}.

Videos in the LLP have 11,849 video-level event annotations on the presence or absence of different video events for facilitating weakly-supervised learning.  
Each video is 10$s$ long and has at least 1$s$ audio or visual events. 
There are 7,202 videos that contain events from more than one event categories and per video has averaged 1.64 different event categories. To evaluate audio-visual scene parsing performance, we annotate individual audio and visual events with second-wise temporal boundaries for randomly selected 1,849 videos from the LLP dataset. Note that the audio-visual event labels can be derived from the audio and visual event labels. Finally, we have totally 6,626 event annotations, including 4,131 audio events and 2,495 visual events for the 1,849 videos. Merging the individual audio and visual labels, we obtain 2,488 audio-visual event annotations. 
To do validation and testing, we split the subset into a validation set with 649 videos and a testing set with 1,200 videos. Our weakly-supervised audio-visual video parsing network will be trained using the 10,000 videos with weak labels and the trained models are developed and tested on the validation and testing sets with fully annotated labels, respectively.

\section{Audio-Visual Video Parsing with Weak Labels}
\label{sec:avsp_problem}

We define the \textit{Audio-Visual Video Parsing} as a task to {group video segments and parse a video into different temporal audio, visual, and audio-visual events associated with semantic labels}. Since event boundary in the LLP dataset was annotated at second-level, video events will be parsed at scene-level not object/instance level in our experimental setting. Concretely, given a video sequence containing both audio and visual tracks, we divide it into $T$ non-overlapping audio and visual snippet pairs $\{V_t, A_t\}_{t=1}^{T}$, where each snippet is 1$s$ long and $V_t$ and $A_t$ denote visual and audio content in the same video snippet, respectively. Let $\textbf{\textit{y}}_t = \{(y_{t}^a, y_{t}^v, y_{t}^{av})|[y_{a}^t]_{c}, [y_{v}^t]_{c}, [y_{av}^t]_{c} \in \{0, 1\}, c = 1, ..., C\}$ be the event label set for the video snippet $\{V_t, A_t\}$, where $c$ refers to the $c$-th event category and $y_{t}^a$, $y_{t}^v$, and $y_{t}^{av}$ denote audio, visual, and audio-visual event labels, respectively. Here, we have a relation: $y_{t}^{av} = y_{t}^{a}*y_{t}^{v}$, which means that audio-visual events occur only when there exists both audio and visual events at the same time and from the same event categories. 

In this work, we explore the audio-visual video parsing in a weakly-supervised manner. 
We only have video-level labels for training, but will predict precise event label sets for all video snippets during testing, which makes the weakly-supervised audio-visual video parsing be a multi-modal multiple instance learning (MMIL) problem. 
Let a video sequence with $T$ audio and visual snippet pairs be a bag. 
Unlike the previous audio-visual event localization~\cite{tian2018audio} that is formulated as a MIL problem~\cite{maron1998framework} where an audio-visual snippet pair is regarded as an instance, each audio snippet and the corresponding visual snippet occurred at the same time denote two individual instances in our MMIL problem. 
So, a positive bag containing video events will have at least one positive video snippet; meanwhile at least one modality has video events in the positive video snippet. 
During training, we can only access bag labels. 
During inference, we need to know not only which video snippets have video events but also which sensory modalities perceive the events. The temporal and multi-modal uncertainty in this MMIL problem makes it very challenging.

\begin{figure}[tb]
    \centering
    \includegraphics[width=\columnwidth]{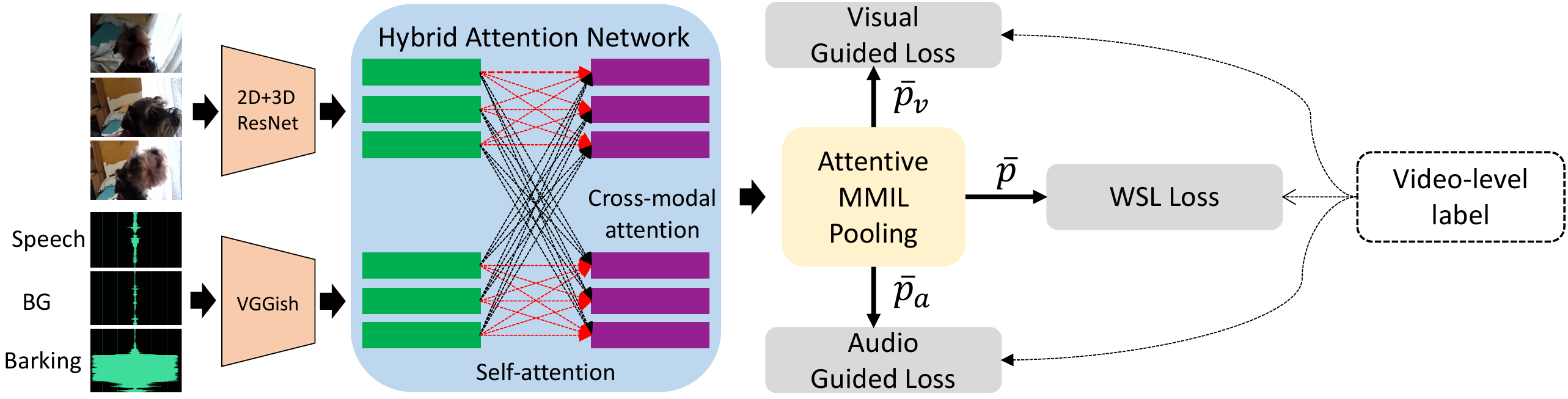}
    \caption{The proposed audio-visual video parsing framework. It uses pre-trained CNNs to extract snippet-level audio and visual features and leverages multimodal temporal contexts with the proposed hybrid attention network (HAN). For each snippet, we will predict both audio and visual event labels from the aggregated features by the HAN. Attentive MMIL pooling is utilized to adaptively predict video-level event labels for weakly-supervised learning (WSL) and individual guided learning is devised to mitigate the modality bias issue.}
    \label{fig:framework}
\end{figure} 

\section{Method}
\label{sec:method}

First, we present the overall framework that formulates the weakly-supervised audio-visual video parsing as an MMIL problem in Sec.~\ref{sec:overview}. 
Built upon this framework, we propose a new multimodal temporal model: hybrid attention network in Sec.~\ref{sec:han};  attentive MMIL pooling in Sec.~\ref{sec:att_pool}; addressing modality bias and noisy label issues in Sec.~\ref{sec:GL_LN}.

\subsection{Audio-Visual Video Parsing Framework}
\label{sec:overview}

Our framework, as illustrated in Fig.~\ref{fig:framework}, has three main modules: audio and visual feature extraction, multimodal temporal modeling, and attentive MMIL pooling. 
Given a video sequence with $T$ audio and visual snippet pairs $\{V_t, A_t\}_{t=1}^{T}$, we first use pre-trained visual and audio models to extract snippet-level visual features: $\{f_v^t\}_{t=1}^{T}$ and audio features: $\{f_a^t\}_{t=1}^{T}$, respectively. 
Taking extracted audio and visual features as inputs, we use two hybrid attention networks as the multimodal temporal modeling module to leverage unimodal and cross-modal temporal contexts and obtain updated visual features $\{\hat{f}_v^t\}_{t=1}^{T}$ and audio features $\{\hat{f}_a^t\}_{t=1}^{T}$. 
To predict audio and visual instance-level labels and make use of the video-level weak labels, we address the MMIL problem with a novel attentive MMIL pooling module outputting video-level labels. 

\subsection{Hybrid Attention Network}
\label{sec:han}

Natural videos tend to contain continuous and repetitive rather than isolated audio and visual content. In particular, audio or visual events in a video usually redundantly recur many times inside the video, both within the same modality (unimodal temporal recurrence~\cite{naphade2002discovering,roma2013recurrence}), as well as across different modalities (audio-visual temporal synchronization~\cite{korbar2018cooperative} and asynchrony~\cite{vroomen2004recalibration}). The observation suggests us to jointly model the temporal recurrence, co-occurrence, and asynchrony in a unified approach. However, existing audio-visual learning methods~\cite{lin2019dual,tian2018audio,wu2019DAM} usually ignore the audio-visual temporal asynchrony and explore unimodal temporal recurrence using temporal models (\emph{e.g.}, LSTM~\cite{hochreiter1997long} and Transformer~\cite{vaswani2017attention}) and audio-visual temporal synchronization using multimodal fusion (\emph{e.g.}, feature fusion~\cite{tian2018audio} and prediction ensemble~\cite{kazakos2019epic}) in a isolated way. To simultaneously capture multimodal temporal contexts, we propose a new temporal model: Hybrid Attention Network (HAN), which uses a self-attention network and a cross-attention network to adaptively learn which bimodal and cross-modal snippets to look for each audio or visual snippet, respectively. 

At each time step $t$, a hybrid attention function $g$ in HAN will be learned from audio and visual features: $\{f_a^{t}, f_v^t\}_{t=1}^{T}$ to update $f_a^{t}$ and $f_v^{t}$, respectively. The updated audio feature $\hat{f}_a^{t}$ and visual feature $\hat{f}_v^{t}$ can be computed as:
\begin{align}
    \hat{f}_a^{t} = g(f_a^{t}, \textbf{\textit{f}}_a, \textbf{\textit{f}}_v) = f_a^t + g_{sa}(f_a^{t}, \textbf{\textit{f}}_a) + g_{ca}(f_a^{t}, \textbf{\textit{f}}_v)\enspace,\\
    \hat{f}_v^{t} = g(f_v^{t}, \textbf{\textit{f}}_a, \textbf{\textit{f}}_v) = f_v^t + g_{sa}(f_v^{t}, \textbf{\textit{f}}_v) + g_{ca}(f_v^{t}, \textbf{\textit{f}}_a)\enspace,
\end{align}
where $\textbf{\textit{f}}_a = [f_a^1;...;f_a^T]$ and $\textbf{\textit{f}}_v= [f_v^1;...;f_v^T]$;  $g_{sa}$ and $g_{ca}$ are self-attention and cross-modal attention functions, respectively; skip-connections can help preserve the identity information from the input sequences. The two attention functions are formulated with the same computation mechanism. With $g_{sa}(f_a^{t}, \textbf{\textit{f}}_a)$ and $g_{ca}(f_a^{t}, \textbf{\textit{f}}_v)$ as examples, they are defined as:
\begin{align}
    g_{sa}(f_a^{t}, \textbf{\textit{f}}_a) = \sum_{t=1}^{T}w_t^{sa}f_a^t = \textit{softmax}(\frac{f_a^t\textbf{\textit{f}}_a^{'}}{\sqrt{d}})\textbf{\textit{f}}_a\enspace,\\
    g_{ca}(f_a^{t}, \textbf{\textit{f}}_v) = \sum_{t=1}^{T}w_t^{ca}f_v^t = \textit{softmax}(\frac{f_a^t\textbf{\textit{f}}_v^{'}}{\sqrt{d}})\textbf{\textit{f}}_v\enspace,
\end{align}
where the scaling factor $d$ is equal to the audio/visual feature dimension and $(\cdot)^{'}$ denotes the transpose operator. Clearly, the self-attention and cross-modal attention functions in HAN will assign large weights to snippets, which are similar to the query snippet containing the same video events within the same modality and cross different modalities. The experimental results show that the HAN modeling unimodal temporal recurrence, multimodal temporal co-occurrence, and audio-visual temporal asynchrony can well capture unimodal and cross-modal temporal contexts and improves audio-visual video parsing performance.

\begin{figure}[tb]
    \centering
    \includegraphics[width=\columnwidth]{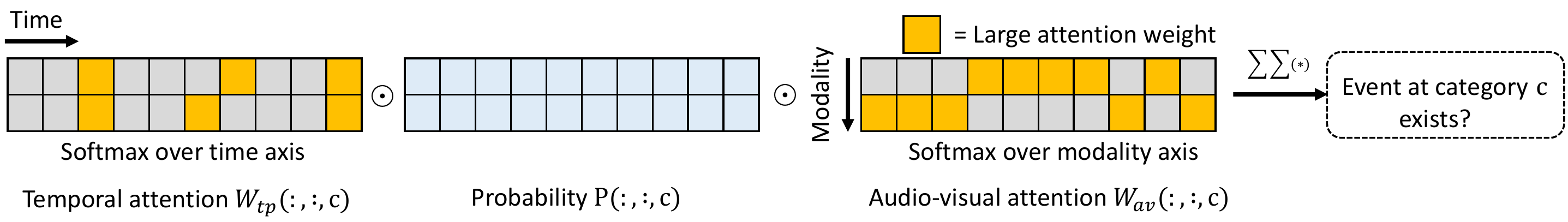}
    \caption{Attentive MMIL Pooling. For event category $c$, temporal and audio-visual attention mechanisms will adaptively select informative event predictions crossing temporal and modality axes, respectively, for predicting whether there is an event at the category. }
    \label{fig:attentive}
\end{figure} 

\subsection{Attentive MMIL Pooling}
\label{sec:att_pool}

To achieve audio-visual video parsing, we predict all event labels for audio and visual snippets from temporal aggregated features: $\{\hat{f}_a^{t}, \hat{f}_v^t\}_{t=1}^{T}$. We use a shared fully-connected layer to project audio and visual features to different event label space and adopt a sigmoid function to output probability for each event category:
\begin{align}
    p_a^t &= sigmoid(FC(\hat{f}_a^{t}))\enspace, \\  p_v^t &= sigmoid(FC(\hat{f}_v^{t}))\enspace,
\end{align}
where $p_a^t$ and $p_v^t$ are predicted audio and visual event probabilities at timestep $t$, respectively. Here, the shared FC layer can implicitly enforce audio and visual features into a similar latent space. The reason to use sigmoid to output an event probability for each event category rather than softmax to predict a probability distribution over all categories is that a single snippet may have multiple labels rather than only a single event as assumed in Tian \emph{et al.}~\cite{tian2018audio}.


Since audio-visual events only occur when sound sources are visible and their sounds are audible, the audio-visual event probability $ p_{av}^{t}$ can be derived from individual audio and visual predictions: $p_{av}^{t} = p_a^t * p_v^t$.
If we have direct supervisions for all audio and visual snippets from different time steps, we can simply learn the audio-visual video parsing network in a fully-supervised manner. However, in this MMIL problem, we can only access a video-level weak label $\bar{\textbf{\textit{{y}}}}$ for all audio and visual snippets: $\{A_t, V_t\}_{t=1}^{T}$ from a video. To learn our network with weak labels, as illustrated in Fig.~\ref{fig:attentive}, we propose a attentive MMIL pooling method to predict video-level event probability: $\bar{\textbf{\textit{p}}}$ from $\{{p}_a^{t}, {p}_v^t\}_{t=1}^{T}$.
Concretely, the $\bar{\textbf{\textit{p}}}$ is computed by:
\begin{align}
\bar{\textbf{\textit{p}}} = \sum_{t=1}^{T}\sum_{m=1}^{M} (W_{tp}\odot W_{av}\odot P)[t, m, :]\enspace,
\end{align}
where $\odot$ denotes element-wise multiplication; $m$ is a modality index and $M$ = $2$ refers to audio and visual modalities; $W_{tp}$ and $W_{av}$ are temporal attention and audio-visual attention tensors predicted from $\{\hat{f}_a^{t}, \hat{f}_v^t\}_{t=1}^{T}$, respectively, and $P$ is the probability tensor built by $\{{p}_a^{t}, {p}_v^t\}_{t=1}^{T}$ and we have $P(t, 0, :) = p_{a}^{t}$ and $P(t, 1, :) = p_{v}^{t}$. To compute the two attention tensors, we first compose an input feature tensor $F$, where  $F(t, 0, :) = \hat{f}_{a}^{t}$ and $F(t, 1, :) = \hat{f}_{v}^{t}$. Then, two different FC layers are used to transform the $F$ into two tensors: $F_{tp}$ and $F_{av}$, which has the same size as $P$. To adpatively select most informative snippets for predicting probabilities of different event categories, we assign different weights to snippets at different time steps with a temporal attention mechanism:
\begin{align}
    W_{tp}[:, m, c] = softmax(F_{tp}[:, m, c])\enspace,
\end{align}
where $m$ = $1, 2$ and $c$ = $1, \dots, C$. Accordingly, we can adaptively select most informative modalities with the audio-visual attention tensor: 
\begin{align}
    W_{av}[t, :, c] = softmax(F_{av}[t, :, c])\enspace,
\end{align}
where $t$ = $1, \dots, T$ and $c$ = $1, \dots, C$. The snippets within a video from different temporal steps and different modalities may have different video events. The proposed attentive MMIL pooling can well model this observation with the tensorized temporal and multimodal attention mechanisms.

With the predicted video-level event probability $\bar{\textbf{\textit{p}}}$ and the ground truth label $\bar{\textbf{\textit{{y}}}}$, we can optimize the proposed weakly-supervised learning model with a binary cross-entropy loss function: $\mathcal{L}_{wsl} = CE(\bar{\textbf{\textit{p}}}, \bar{\textbf{\textit{y}}}) = -\sum_{c=1}^{C} \bar{\textbf{\textit{y}}}[c]log(\bar{\textbf{\textit{p}}}[c])$.

\subsection{Alleviating Modality Bias and Label Noise}
\label{sec:GL_LN}

The weakly supervised audio-visual video parsing framework only uses less detailed annotations without requiring expensive densely labeled audio and visual events for all snippets. This advantage makes this weakly supervised learning framework appealing. However, it usually enforces models to only identify discriminative patterns in the training data, which was observed in previous weakly-supervised MIL problems~\cite{singh2017hide,song2014weakly,zhou2016learning}. In our MMIL problem, the issue becomes even more complicated since there are multiple modalities and different modalities might not contain equally discriminative information. With weakly-supervised learning, the model tends to only use information from the most discriminative modality but ignore another modality, which can probably achieve good video classification performance but terrible video parsing performance on the events from ignored modality and audio-visual events. Since a video-level label contains all event categories from audio and visual content within the video, to alleviate the modality bias in the MMIL, we propose to use explicit supervisions to both modalities with a guided loss:
\begin{align}
\label{eq:gl}
    \mathcal{L}_{g} = CE(\bar{\textbf{\textit{p}}}_a, \bar{\textbf{\textit{y}}}_a) + CE(\bar{\textbf{\textit{p}}}_v, \bar{\textbf{\textit{y}}}_v)\enspace,
\end{align}
where $\bar{\textbf{\textit{y}}}_a = \bar{\textbf{\textit{y}}}_v = \bar{\textbf{\textit{y}}}$, and $\bar{\textbf{\textit{p}}}_a = \sum_{t=1}^{T}(W_{tp}\odot P)[t, 0, :]$ and $\bar{\textbf{\textit{p}}}_v = \sum_{t=1}^{T}(W_{tp}\odot P)[t, 1, :]$ are video-level audio and visual event probabilities, respectively. 

However, not all video events are audio-visual events, which means that an event occurred in one modality might not occur in another modality and then the corresponding event label will be label noise for one of the two modalities. Thus, the guided loss: $\mathcal{L}_{g}$ suffers from noisy label issue. For the example shown in Fig.~\ref{fig:framework}, the video-level label is \{\textit{Speech}, \textit{Dog}\} and the video-level visual event label is only \{\textit{Dog}\}. The \{\textit{Speech}\} will be a noisy label for the visual guided loss. 
To handle the problem, we use label smoothing~\cite{szegedy2016rethinking} to lower the confidence of positive labels with
smoothing $\bar{\textbf{\textit{y}}}$ and generate smoothed labels:  $\bar{\textbf{\textit{y}}}_a$ and $\bar{\textbf{\textit{y}}}_v$. They are formulated as: $\bar{\textbf{\textit{y}}}_a  = (1 - \epsilon_a) \bar{\textbf{\textit{y}}} + \frac{\epsilon_a}{K}$ and $\bar{\textbf{\textit{y}}}_v  = (1 - \epsilon_v) \bar{\textbf{\textit{y}}} + \frac{\epsilon_v}{K}$,
where $\epsilon_a, \epsilon_v\in [0,1)$ are two confidence parameters to balance the event probability distribution and a uniform distribution: $u = \frac{1}{K}$ ($K > 1$).
For a noisy label at event category $c$, when $\bar{\textbf{\textit{y}}}[c] = 1$ and real $\bar{\textbf{\textit{y}}}_a[c] = 0$, we have $\bar{\textbf{\textit{y}}}[c] = (1 - \epsilon_a) \bar{\textbf{\textit{y}}}[c] + \epsilon_a > (1 - \epsilon_a) \bar{\textbf{\textit{y}}}[c] + \frac{\epsilon_a}{K}=\bar{\textbf{\textit{y}}}_a[c]$ and the smoothed label will become more reliable. Label smoothing technique is commonly adopted in a lot of tasks, such as image classification~\cite{szegedy2016rethinking}, speech recognition~\cite{chorowski2016towards}, and machine translation~\cite{vaswani2017attention} to reduce over-fitting and improve generalization capability of deep models. Different from the past methods, we use smoothed labels to mitigate label noise occurred in the individual guided learning. Our final model is optimized with the two loss terms: $\mathcal{L} = \mathcal{L}_{wsl} + \mathcal{L}_{g}$.

\section{Experiments}

\subsection{Experimental Settings}
\label{base&metric}

\noindent \textbf{Implementation Details.} For a 10-second-long video, we first sample video frames at $8 fps$ and each video is divided into non-overlapping snippets of the same length with $8$ frames in 1 second. Given a visual snippet, we extract a $512$-D snippet-level feature with fused features extracted from ResNet152~\cite{he2016deep} and 3D ResNet~\cite{tran2018closer}.  In our experiments, batch size and number of epochs are set as 16 and 40, respectively. The initial learning rate is $3e$-$4$ and will drop by multiplying $0.1$ after every 10 epochs. Our models optimized by Adam can be trained using one NVIDIA 1080Ti GPU.

\noindent \textbf{Baselines.} Since there are no existing methods to address the audio-visual video parsing, we design several baselines based on previous state-of-the-art weakly-supervised sound detection~\cite{kong2018audio,wang2019comparison}, temporal action localization~\cite{liu2019completeness,nguyen2018weakly}, and audio-visual event localization~\cite{lin2019dual,tian2018audio} methods to validate the proposed framework. To make \cite{lin2019dual,tian2018audio} possible to address audio-visual scene parsing, we add additional audio and visual branches to predict audio and visual event probabilities supervised with an additional guided loss as defined in Sec.~\ref{sec:GL_LN}. For fair comparisons, the compared approaches use the same audio and visual features as our method.

\noindent \textbf{Evaluation Metrics.} To comprehensively measure the performance of different methods, we evaluate them on parsing all types of events (individual audio, visual, and audio-visual events) under both segment-level and event-level metrics. To evaluate overall audio-visual scene parsing performance, we also compute aggregated results, where Type@AV computes averaged audio, visual, and audio-visual event evaluation results and Event@AV computes the F-score considering all audio and visual events for each sample rather than directly averaging results from different event types as the Type@AV. We use both segment-level and event-level F-scores~\cite{mesaros2016metrics} as metrics. The segment-level metric can evaluate snippet-wise event labeling performance. For computing event-level F-score results, we extract events with concatenating positive consecutive snippets in the same event categories and compute the event-level F-score based on {mIoU} = 0.5 as the threshold. 

\setlength{\tabcolsep}{3pt}
\begin{table}[t]
\begin{center}
\caption{Audio-visual video parsing accuracy ($\%$) of different methods on the LLP test dataset. These methods all use the same audio and visual features as inputs for a fair comparison. The top-1 results in each line are highlighted.}
\label{tbl:avsp}
\scalebox{1.0}{
\begin{tabular}{l| c| c  c }
\toprule
Event type&Methods &Segment-level  &Event-level \\
\midrule
\multirow{6}{*}{Audio}
    &Kong \emph{et. al} 2018~\cite{kong2018audio}&39.6&29.1\\
        &TALNet~\cite{wang2019comparison}&{50.0}&{41.7}\\
        &AVE~\cite{tian2018audio}&47.2&40.4\\
    &AVSDN~\cite{lin2019dual}&47.8&34.1\\
 &Ours& \textbf{60.1}&\textbf{51.3}\\
\midrule
\multirow{5}{*}{Visual}
   &STPN~\cite{nguyen2018weakly} &46.5&41.5\\
   &CMCS~\cite{liu2019completeness}&48.1&45.1\\
   &AVE~\cite{tian2018audio}&37.1&34.7\\
   &AVSDN~\cite{lin2019dual}&52.0&46.3\\
 &Ours&\textbf{52.9}&\textbf{48.9}\\
 \midrule
\multirow{3}{*}{Audio-Visual}&AVE~\cite{tian2018audio}&35.4&31.6\\
    &AVSDN~\cite{lin2019dual}&37.1&26.5\\
 &Ours&\textbf{48.9}&\textbf{43.0}\\
 \midrule
 \multirow{3}{*}{Type@AV}&AVE~\cite{tian2018audio}&39.9&35.5\\
    &AVSDN~\cite{lin2019dual}&45.7&35.6\\
 &Ours&\textbf{54.0}&\textbf{47.7}\\
 \midrule
 \multirow{3}{*}{Event@AV}&AVE~\cite{tian2018audio}&41.6&36.5\\
    &AVSDN~\cite{lin2019dual}&50.8&37.7\\
 &Ours&\textbf{55.4}&\textbf{48.0}\\
\bottomrule
\end{tabular}
}
\end{center}
\end{table}

\subsection{Experimental Comparison}
\label{sec:comp}

To validate the effectiveness of the proposed audio-visual video parsing network, we compare it with weakly-supervised sound event detection methods: Kong \emph{et al} 2018~\cite{kong2018audio} and TALNet~\cite{wang2019comparison} on audio event parsing, weakly-supervised action localization methods: STPN~\cite{nguyen2018weakly} and  CMCS~\cite{liu2019completeness} on visual event parsing, and modified audio-visual event localization methods: AVE~\cite{tian2018audio} and AVSD~\cite{lin2019dual} on audio, visual, and audio-visual event parsing. The quantitative results are shown in Tab.~\ref{tbl:avsp}. We can see that our method outperforms compared approaches on all audio-visual video parsing subtasks under both the segment-level and event-level metrics, which demonstrates that our network can predict more accurate snippet-wise event categories with more precise event onsets and offsets for testing videos. 
\setlength{\tabcolsep}{2pt}
\begin{table}[t]
\begin{center}
\caption{Ablation study on learning mechanism, attentive MMIL pooling, hybrid attention network, and handling noisy labels. Segment-level audio-visual video parsing results are shown. The best results for each ablation study are highlighted.}
\label{tbl:ablation}
\scalebox{0.715}{
\begin{tabular}{c| c| c| c |c  c c| c c}
\toprule
Loss &MMIL Pooling& Temporal Net &Handle Noisy Label &Audio &Visual  &Audio-Visual&Type@AV &Event@AV\\
\midrule
\textcolor{blue}{$\mathcal{L}_{wsl}$}& Attentive&$\times$&$\times$&\textbf{56.9}& 16.4 &17.2 &30.2&43.3\\
\textcolor{blue}{$\mathcal{L}_g$}& Attentive&$\times$&$\times$& 42.3& 43.9& 34.5&40.3&42.0\\
 \textcolor{blue}{$\mathcal{L}_{wsl} + \mathcal{L}_g$}& Attentive&$\times$&$\times$&45.1& \textbf{51.7} &\textbf{35.0} &\textbf{44.0} & \textbf{48.9}\\
 \midrule
 $\mathcal{L}_{wsl} + \mathcal{L}_g$& \textcolor{blue}{Max}&$\times$ &$\times$&31.6&43.6&22.5&32.6&39.1\\
 $\mathcal{L}_{wsl} + \mathcal{L}_g$& \textcolor{blue}{Mean}&$\times$&$\times$& 40.2&43.2&\textbf{35.0}&39.5&39.7\\
 $\mathcal{L}_{wsl} + \mathcal{L}_g$& \textcolor{blue}{Attentive}&$\times$&$\times$&\textbf{45.1}& \textbf{51.7} &\textbf{35.0} &\textbf{44.0} & \textbf{48.9}\\
  \midrule
   $\mathcal{L}_{wsl} + \mathcal{L}_g$& Attentive&\textcolor{blue}{$\times$}&{$\times$}&45.1& {51.7} &{35.0} &{44.0} & {48.9}\\
 $\mathcal{L}_{wsl} + \mathcal{L}_g$& Attentive&\textcolor{blue}{GRU}~\cite{cho2014learning}&$\times$&52.0&49.4&39.0&46.8&51.0\\
 $\mathcal{L}_{wsl} + \mathcal{L}_g$& Attentive&\textcolor{blue}{Transformer}~\cite{vaswani2017attention}&$\times$&53.4& \textbf{53.8}& 41.8&49.7&53.3\\
 $\mathcal{L}_{wsl} + \mathcal{L}_g$& Attentive&\textcolor{blue}{HAN}&$\times$&\textbf{58.4}& {52.8}&\textbf{48.4}&\textbf{53.2}&\textbf{54.5}\\
 \midrule
\textcolor{blue}{$\mathcal{L}_{wsl}$}& Attentive&HAN&$\times$&39.6&40.5&20.1&33.4&44.9\\
\textcolor{blue}{$\mathcal{L}_g$}& Attentive&HAN&$\times$& 57.5&52.5&47.4&52.5&53.8\\
 \textcolor{blue}{$\mathcal{L}_{wsl} + \mathcal{L}_g$}& Attentive&HAN&$\times$&\textbf{58.4}& \textbf{52.8}&\textbf{48.4}&\textbf{53.2}&\textbf{54.5}\\
 \midrule
 $\mathcal{L}_{wsl} + \mathcal{L}_g$& \textcolor{blue}{Max}&HAN &$\times$&55.7&52.0&\textbf{48.6}&52.1&51.8\\
 $\mathcal{L}_{wsl} + \mathcal{L}_g$& \textcolor{blue}{Mean}&HAN&$\times$& 56.0&51.9&46.3&51.4&52.9\\
 $\mathcal{L}_{wsl} + \mathcal{L}_g$& \textcolor{blue}{Attentive}&HAN&$\times$&\textbf{58.4}& \textbf{52.8}&{48.4}&\textbf{53.2}&\textbf{54.5}\\
   \midrule
    $\mathcal{L}_{wsl} + \mathcal{L}_g$& Attentive&HAN&\textcolor{blue}{$\times$}&{58.4}& {52.8}&{48.4}&{53.2}&{54.5}\\
 $\mathcal{L}_{wsl} + \mathcal{L}_g$& Attentive&HAN&\textcolor{blue}{Bootstrap}~\cite{reed2014training}&59.0&52.6&47.8&53.1&55.2\\
 $\mathcal{L}_{wsl} + \mathcal{L}_g$& Attentive&HAN&\textcolor{blue}{Label Smoothing~\cite{szegedy2016rethinking}}&\textbf{60.1}& \textbf{52.9}&\textbf{48.9}&\textbf{54.0}&\textbf{55.4}\\
 \bottomrule
\end{tabular}
}
\end{center}
\end{table}

\noindent \textbf{Individual Guided Learning.} From Tab.~\ref{tbl:ablation}, we observe that the model without individual guided learning can achieve pretty good performance on audio event parsing but incredibly bad visual parsing results leading to terrible audio-visual event parsing; w/ only $\mathcal{L}_{g}$ model can achieve both reasonable audio and visual event parsing results; our model trained with both $\mathcal{L}_{wsl}$ and $\mathcal{L}_{g}$ outperforms model train without and with only $\mathcal{L}_{g}$. The results indicate that the model trained only $\mathcal{L}_{wsl}$ find discriminative information from mostly sounds and visual information is not well-explored during training and the individual learning can effectively handle the modality bias issue. In addition, when the network is trained with only $\mathcal{L}_{g}$, it actually models audio and visual event parsing as two individual MIL problems in which only noisy labels are used. Our MMIL framework can learn from clean weak labels with $\mathcal{L}_{wsl}$ and handle the modality bias with $\mathcal{L}_{g}$ achieves the best overall audio-visual video parsing performance. 
Moreover, we would like to note that the modality bias issue is from audio and visual data unbalance in training videos, which are originally from an audio-oriented dataset: AudioSet. Since the issue occurred after just 1 epoch training, it is not over-fitting.

\noindent \textbf{Attentive MMIL Pooling.} To validate the proposed Attentive MMIL Pooling, we compare it with two commonly used methods: Max pooling and Mean pooling. Our Attentive MMIL Pooling (see Tab.~\ref{tbl:ablation}) is superior over the both compared methods. The Max MMIL pooling only selects the most discriminative snippet for each training video, thus it cannot make full use of informative audio and visual content. The Mean pooling does not distinguish the importance of different audio and visual snippets and equally aggregates instance scores in a bad way, which can obtain good audio-visual event parsing but poor individual audio and visual event parsing since a lot of audio-only and visual-only events are incorrectly parsed as audio-visual events.
Our attentive MMIL pooling allows assigning different weights to audio and visual snippets within a video bag for each event category, thus can adaptively discover useful snippets and modalities.

\noindent \textbf{Hybrid Attention Network.} We compare our HAN with two popular temporal networks: GRU and Transformer and a base model without temporal modeling in Tab. \ref{tbl:ablation}. The models with GRU and Transformer are better than the base model and our HAN outperforms the GRU and Transformer. The results demonstrate that temporal aggregation with exploiting temporal recurrence is important for audio-visual video parsing and our HAN with jointly exploring unimodal temporal recurrence, multimodal temporal co-occurrence, and audio-visual temporal asynchrony is more effective in leveraging the multimodal temporal contexts. Another surprising finding of the HAN is that it actually tends to alleviate the modality bias by enforcing cross-modal modeling.

\noindent \textbf{Noisy Label.} Tab.~\ref{tbl:ablation} also shows results of our model without handling the noisy label, with Bootstrap~\cite{reed2014training} and label smoothing-based method. We can find that Bootstrap updating labels using event predictions even decreases performance due to error propagation. Label smoothing-based method with reducing confidence for potential false positive labels can help to learn a more robust model with improved audio-visual video parsing results.

\section{Limitation}

To mitigate the modality bias issue, the guided loss is introduced to enforce that each modality should also be able to make the correct prediction on its own. Then, a new problem appears: the guide loss is not theoretically correct because some of the events only appear in one modality, so the labels are wrong. Finally, the label smoothing is used to alleviate the label noise. Although the proposed methods work at each step, they also introduce new problems. It is worth to design a one-pass approach. One possible solution is to introduce a new learning strategy to address the modality bias problem rather than using the guided loss. For example, we could perform modality dropout to enforce the model to explore both audio and visual information during training.



\section{Conclusion and Future Work}

In this work, we investigate a fundamental audio-visual research problem: audio-visual video parsing in a weakly-supervised manner. We introduce baselines and propose novel algorithms to address the problem.  Extensive experiments on the newly collected LLP dataset support our findings that the audio-visual video parsing is tractable even learning from cheap weak labels, and the proposed model is capable of leveraging multimodal temporal contexts, dealing with modality bias, and mitigating label noise. 
Accurate audio-visual video parsing opens the door to a wide spectrum of potential applications, as discussed below.

\begin{figure}[tb]%
    \centering
    \subfloat[Asynchronous Separation]{{\includegraphics[width=0.36\columnwidth]{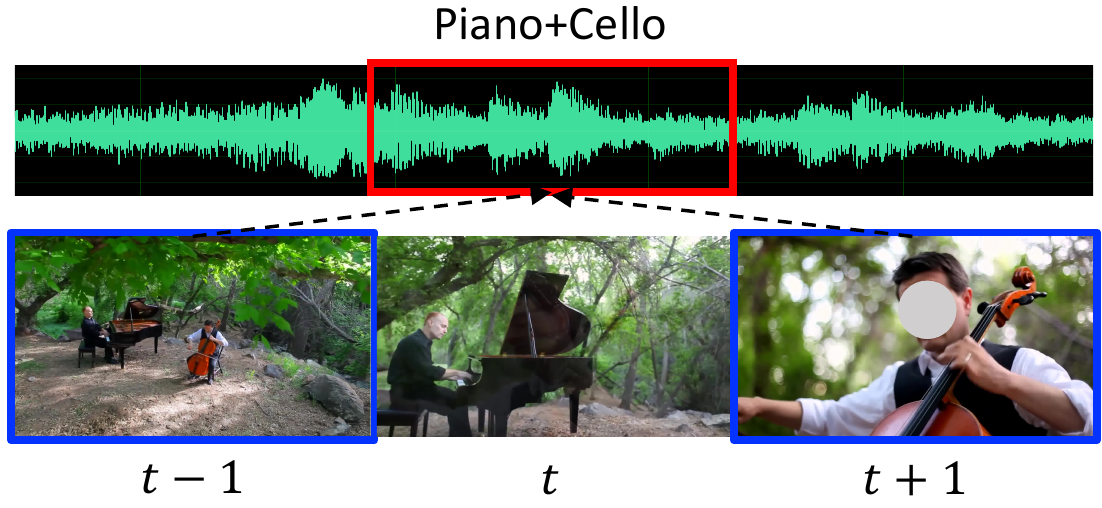} }}%
    \subfloat[Scene-Aware Video Understanding]{{\includegraphics[width=0.6\columnwidth]{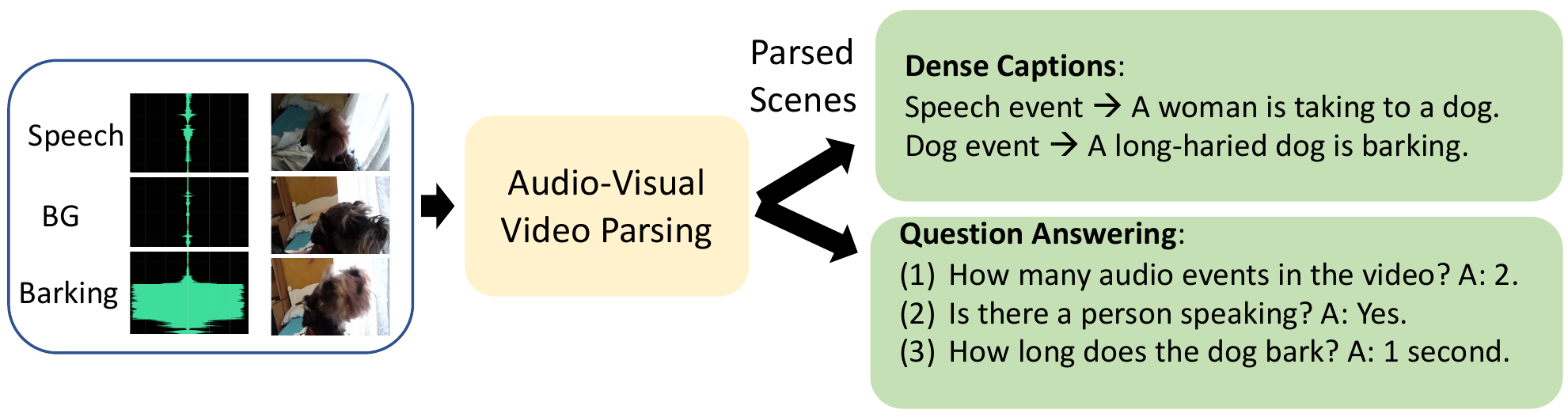} }}%
    \caption{Potential applications of audio-visual video parsing. (a) Temporally asynchronous visual events detected by audio-visual video parsing highlighted in blue boxes can provide related visual information to separate \textit{Cello} sound from the audio mixture in the red box. (b) Parsed scenes can provide important cues for audio-visual scene-aware video dense captioning and question answering.}%
     \label{fig:app}
\end{figure}


\noindent{\textbf{Asynchronous Audio-Visual Sound Separation.}} Audio-visual sound separation approaches use sound sources in videos as conditions to separate the visually indicated individual sounds from sound mixtures~\cite{ephrat2018looking,gao2018learning,gao20192,gao2019co,zhao2019sound,zhao2018sound}. The underlying assumption is that sound sources are visible. However, sounding objects can be occluded or not recorded in videos and the existing methods will fail to handle these cases. Our audio-visual video parsing model can find temporally asynchronous cross-modal events, which can help to alleviate the problem. For the example in Fig.~\ref{fig:app} (a), the existing audio-visual separation models will fail to separate the \textit{Cello} sound from the audio mixture at the time step $t$, since the sound source \textit{Cello} is not visible in the segment. However, our model can help to find temporally asynchronous visual events with the same semantic label as the audio event \textit{Cello} for separating the sound. In this way, we can improve the robustness of audio-visual sound separation by leveraging temporally asynchronous visual content identified by our audio-visual video parsing models.

\noindent{\textbf{Audio-Visual Scene-Aware Video Understanding.}}  The current video understanding community usually focuses on the visual modality and regards information from sounds as a bonus assuming that audio content should be associated with the corresponding visual content. However, we want to argue that auditory and visual modalities are equally important and most natural videos contain numerous audio, visual, and audio-visual events rather than only visual and audio-visual events. 
Our audio-visual scene parsing can achieve a unified multisensory perception, therefore it has the potential to help us build an audio-visual scene-aware video understanding system regarding all audio and visual events in videos(see Figure~\ref{fig:app} (b)).

\noindent \textbf{Acknowledgment}
We thank the anonymous reviewers for the constructive feedback. This work was supported in part by NSF 1741472, 1813709, and 1909912. The article solely reflects the opinions and conclusions of its authors but not the funding agents.

\clearpage
%
%
\bibliographystyle{splncs04}
\bibliography{egbib}
\end{document}